# Regularized Maximum Likelihood for Intrinsic Dimension Estimation

**Mithun Das Gupta**
Epson Research and Development
San Jose, CA
mdasgupta@erd.epson.com

**Thomas S. Huang**
University of Illinois,
Urbana Champaign
huang@ifp.uiuc.edu

## Abstract

We propose a new method for estimating the intrinsic dimension of a dataset by applying the principle of regularized maximum likelihood to the distances between close neighbors. We propose a regularization scheme which is motivated by divergence minimization principles. We derive the estimator by a Poisson process approximation, argue about its convergence properties and apply it to a number of simulated and real datasets. We also show it has the best overall performance compared with two other intrinsic dimension estimators.

## 1 Introduction

The curse of dimensionality can be greatly alleviated if the intrinsic dimension of the data were a prior knowledge. The intrinsic dimension (ID) of a data set roughly translates to the minimum number of free variables needed to generate the data. The task of large scale data handling, namely classification, regression, retrieval etc., all benefit from the fact that the data points are not truly high-dimensional. Although the data is embedded in a high-dimensional space, it can be efficiently summarized in a space of a much lower dimension, such as a nonlinear manifold. This leads to ID estimation as well as manifold based decompositions as two most valuable pre-processing steps. Traditional methods for dimensionality reduction include principal component analysis (PCA), which only deals with linear projections of the data, and multidimensional scaling (MDS), which aims at preserving pairwise distances. More recently, there has been growing interest in manifold projection methods such as Locally Linear Embedding (LLE) [18], Isomap [24], Laplacian and Hessian Eigenmaps [1, 5], which focus on finding a nonlinear low-dimensional embedding of high-dimensional data. One common limitation of these techniques, is the requirement of the prior knowledge of the ID of the manifold.

The dimension of the embedding is a key parameter for manifold projection methods: if the projected dimension is smaller compared to the ID, important data features are collapsed onto the same dimension, and if the dimension is too large, the projections become noisy and, in some cases, unstable. Though the field of sparse coding with over complete bases [16], tries to answer some of the problems with higher dimensional representations, it does not make ID estimation any less important.

Certain methods (e.g., LLE and the Isomap) also require a scale parameter that determines the size of the local neighborhoods used in the algorithms. In this case, it is useful if the dimension estimate is provided as a function of the scale. Kégl [10] discusses an example where the intrinsic dimension is a function of the resolution. Nearest neighbor searching algorithms can also profit from a good dimension estimate. The complexity of search data structures (e.g., kd-trees and R-trees) increase exponentially with the dimension. It was shown by Chávez *et al.* [3] that the complexity increases with the ID of the data rather than the dimension of the embedding space.

The existing approaches for ID estimation can be broadly divided into two groups: eigen projection techniques, and geometric methods. Eigen projection techniques infer ID from the eigen decomposition of the covariance matrix. The number of eigen values larger than a certain threshold is used as an estimation of the ID [6]. The inherent assumption is that the covariance matrix can be computed.

The geometric methods exploit the intrinsic geometry of the dataset and are most often based on fractal dimensions or nearest neighbor (NN) distances. The most popular fractal dimension, within the intrinsic dimension community, is the correlation dimension [7, 2]: given a set $Sn = \{x_1, \ldots, x_n\}$ in a metric space, define

$$C_n(r) = \frac{2}{n(n-1)} \sum_{i=1}^{n} \sum_{j=i+1}^{n} \mathbf{1}\{\|x_i - x_j\| < r\} \quad (1)$$

The correlation dimension is estimated by plotting $\log C_n(r)$ against $\log r$ and estimating the slope of the lin-ear part [7]. The correlation dimension (Eq. (1)) implicitly uses NN distances, which rely on the following fact: if $x_1, \ldots, x_n$ are $d$-dimensional, independent identically distributed (i.i.d.) samples from a density $f(x)$ residing in $\mathbb{R}^m$, $m$ being the ID of the data ($m \ll d$), and $T_k(x)$ is the Euclidean distance from a fixed point $x$ to its $k^{th}$ NN in the sample, then

$$\frac{k}{n} \approx f(x) v_m [T_k(x)]^m \quad (2)$$

where $v_m = \frac{\pi^{m/2}}{\Gamma(m/2+1)}$ is the volume of the unit sphere in $\mathbb{R}^m$. The above relation simply suggests that the proportion of sample points falling into a ball around $x$ is roughly $f(x)$ times the volume of the ball. The relationship in Eq. (2) can be used to estimate the dimension $m$ by regressing $\log \overline{T}_k$ on $\log k$ over a suitable range of $k$, where $\overline{T}_k = n^{-1} \sum_{i=1}^n T_k(x_i)$ is the average of distances from each point to its $k^{th}$ NN [17, 25].

Levina and Bickel (LB) [12] proposed a Maximum Likelihood (ML) estimator based on a Poisson distribution, which computes the intrinsic dimension at each point. The local estimators are then averaged, under the assumption of a single uniform manifold. The experimental evaluation of their technique leads to an estimator which has poor bias at low neighborhood density, and large variance overall. These features about the LB method were noted by Mackay *et al.* [13]. Our extension of LB method targets at improving the variance of the estimator by incorporating a penalty based on the Kullback-Leibler (KL) divergence [11]. Our penalization scheme draws motivation from the recent array of work concentrating on sample variance penalization methods proposed in [14, 15] as well as covariance estimators for high-dimensional spaces as proposed in [4, 9, 8]. Our approach is based on the intuition that, since the estimate relies on the neighboring data points, hence the divergence of the estimate, from the estimates at the neighbors should be close to each other. Working with a Poisson distribution leads to a very intuitive one sided KL divergence. These intuitions lead to a quadratic equation formulation for the intrinsic dimension estimator.

The structure of the paper is as follows. In Sec. 2, we elaborate on the LB technique, since it is our starting point. Sec. 3 describes the details of our technique, and also argues the asymptotic properties of our estimator. Sec. 4 details the results on synthetic as well as real data sets, and we conclude in Sec. 5.

## 2 A Maximum Likelihood Estimator of Intrinsic Dimension

In this section we lay down the specifics of the maximum likelihood estimator (MLE) of the dimension $m$ as proposed by Levina *et al.* [12]. Assume a cloud of points $X_1, \ldots, X_n$ in $\mathbb{R}^d$ to be i.i.d. observations which represent an embedding of a lower-dimensional sample, i.e., $X_i = g(Y_i)$, where $Y_i$ are sampled from an unknown smooth density $f$ on $\mathbb{R}^m$, with unknown $m \ll d$, and $g$ is a continuous and sufficiently smooth (but not necessarily globally isometric) mapping. This assumption ensures that close neighbors in $\mathbb{R}^m$ are mapped to close neighbors in the embedding.

The basic idea is to fix a point $x$, assume a probability measure $f(x) \geq 0$ to be constant in a small sphere $S_x(r)$ of radius $r$ around $x$, and treat the neighbors within this meghborhood as a Poisson process in $S_x(r)$. Consider the inhomogeneous process $\{N(t,x); 0 \geq t \geq r\}$,

$$N(t,x) = \sum_{i=1}^n \mathbf{1}\{X_i \in S_x(t)\} \quad (3)$$

which counts observations within distance $t$ from $x$. Assume the samples which fall in $S_x(r)$ follow a Bernoulli distribution with probability of success $f(x)V$, where $V = v_m r^m$, is the volume of the sphere of radius $r$ centered at $x$. This is a binomial process and with the assumption that if $n \to \infty$, $k \to \infty$, and $k/n \to 0$, we can approximate it as a Poisson process. Consequently the conditional rate of the Poisson process can be approximated as

$$\lambda(r) = f(x) v_m m r^{m-1} \quad (4)$$

Now we can write the log-likelihood of the observed process as

$$L(m, \theta) = \int_0^R \log \lambda(r) dN(r) - \int_0^R \lambda(r) dr \quad (5)$$

where $\theta \doteq \log f(x)$ is the density parameter, and the first integral is a Riemann-Stieltjes integral [23]. The maximum likelihood estimator satisfies $\partial L/\partial \theta = 0$ and $\partial L/\partial m = 0$, leading to a computation for the local dimension at the point $x$. This formulation has been presented in [12]. The simple formulation leads to the estimate

$$\hat{m}_k(x) = \left[ \frac{1}{k-1} \sum_{j=1}^{k-1} \log \frac{T_k(x)}{T_j(x)} \right]^{-1} \quad (6)$$

The estimate is then averaged over multiple values of $k$,

$$\hat{m}_x = \frac{1}{k_2 - k_1 + 1} \sum_{k=k_1}^{k_2} \hat{m}_k(x) \quad (7)$$

and then averaged over all the data points to generate the final estimate $\hat{m}_k = \frac{1}{n} \sum_{i=1}^n \hat{m}_k(x_i)$. This ML estimate for the intrinsic dimension has been commented upon in [13], which proposes averaging quantity $1/\hat{m}$ to generate the ML estimate. The technique presented in [12] presents one of the first attempts at modeling the estimation of intrinsic dimension as a likelihood maximization problem. One serious drawback of this technique is the multiple levels of averaging required for different $k$ as denoted by Eq. (7). Authors in [13] also point out the high bias of the estimated dimension $m$ even for small $k$. For larger values of $k$ a high bias is expected since the locally smooth density assumption around any data point fails at higher $k$. But for smaller values of $k$ the bias for the estimated $m$ is quite large. One remedy for this problem, proposed in [13], was to estimate the inverse of the dimension, namely $1/m$ to reduce the bias. The results seem to be promising, but they are based on the assumption that the data surrounding each of the n-points are independent, which is clearly incorrect.

## 3 Regularized Maximum Likelihood Estimator

For the exponential family of distributions, it has been shown that the MLE is unbounded [21, 20, 22, 19]. For the inhomogeneous Poisson process mentioned in Eq. (3), the partial log-likelihood function of Eq. (5) becomes unbounded for $m \to 0$. In this paper we propose a regularization scheme which tries to answer some of the problems of the LB scheme.

To facilitate the selection of a regularization function we look at the Kullback-Leibler (KL) divergence [11] between the rate parameters of the Poisson process. Let us assume that the actual process has a rate parameter $\lambda_0$ with associated *true dimension* $m_0$, which is uniform over the neighborhood of $x$ defined by $r$. Now assuming the radius $r$ around $x$ to be reasonably well behaved[1], we can write the directed KL divergence between Poisson($\lambda_x$) and Poisson($\lambda_0$) as

$$\begin{aligned} D_{KL}(\lambda_x \| \lambda_0) &= \lambda_0 - \lambda_x + \lambda_x \log \frac{\lambda_x}{\lambda_0} \quad (8) \\ &= \lambda_0 (1 - \frac{\lambda_x}{\lambda_0} + \frac{\lambda_x}{\lambda_0} \log \frac{\lambda_x}{\lambda_0}) \end{aligned}$$

Close to the optimal point, the ratio $\frac{\lambda_x}{\lambda_0}$ can be further simplified as

$$\frac{\lambda_x}{\lambda_0} = \frac{v_{m_x} m_x r^{m_x - 1}}{v_{m_0} m_0 r^{m_0 - 1}} \approx \frac{m_x}{m_0}$$

where we assume $\frac{v_{m_x}}{v_{m_0}} r^{m_x - m_0} \approx 1$. In essence the divergence term can now be written as

$$D_{KL}(\lambda_x \| \lambda_0) = c D(m_x \| m_0)$$

where

$$D(m_x \| m_0) \doteq m_0 - m_x + m_x \log \frac{m_x}{m_0} \quad (9)$$

and $c$ is a constant. Note that the quantity $D(.)$ does not have the subscript $KL$ to denote that it is a scaled distance metric, similar, but not equal to the KL divergence.

---
[1]The behavior of $r$ has been illustrated delightfully well in [10], Fig.1

Another important aspect of this formulation is the ease of finding the ML estimate of the quantity $m_0$, at least close to optimality. If we accumulate the divergence values around the neighborhood of the point $x$ and differentiate with respect to $m_0$, it can be shown that the value which minimizes the sum of divergences is the arithmetic mean of the current estimates $m_i$ at the neighborhood locations.

$$\begin{aligned} \tfrac{\partial}{\partial m_0} \sum_{y \in S_x(r)} D(m_y \| m_0) &= 0 & (10) \\ \Rightarrow \sum_y (1 - \tfrac{m_y}{m_0}) &= 0 & (11) \\ \Rightarrow k - \tfrac{\sum_y m_y}{m_0} &= 0 & (12) \\ \Rightarrow m_0 = \tfrac{\sum_y m_y}{k} & & (13) \end{aligned}$$

Denoting this estimate by $\bar{m}_0$, leads us to the expression

$$\text{MLE}(m_0)|_{x,r} \equiv \bar{m}_0(x, r) = \frac{1}{k} \sum_{i=1}^{k} m_i \quad (14)$$

where $(x, r)$ signifies that the ML estimate of the true dimension is dependent on the current point $x$ as well its neighborhood parameterized by $r$. To further simplify the notations we drop the notational dependence on $(x, r)$, and write the approximate optimal estimate as $\bar{m}_0$, whenever the context is self explanatory. The estimation of $\bar{m}_0$ and $m_x$ hints at an EM algorithm like scheme, where, in the E step, we average the the current estimates of dimension at the neighbors ($m_i$'s) to generate $\bar{m}_0$, and in the M step update the individual estimates $m_x$'s at all the data locations.

The formulation till this point can now be put into our generic optimization scheme for obtaining the ID estimate at a single point $x$

$$\arg\max_m \int_0^R \log \lambda(r) dN(r) - \int_0^R \lambda(r) dr \quad (15)$$
$$-\gamma D(m \| \bar{m}_0)$$

where the dependence on $x$ is assumed implicit. The boundedness of our estimates is guaranteed by the constraint, and it also ensures that we remain close to the optimal point at all times. Writing the first order derivatives for Eq. (15) with respect to $\theta$ and $m$ and equating them to zero, we get

$$N(R) = e^\theta v_m m R^m \quad (16)$$

$$\frac{N(R)}{m} + \int_0^R \log t \, dN(t) - N(R) \log R \quad (17)$$
$$- \gamma \frac{\partial}{\partial m} D(m \| \bar{m}_0) = 0$$

At this point we transform the problem into a more tangible one, such that the radius $R$ considered is actually $R_k$, the distance of the $k^{th}$ NN from the point $x$, and consequently replace $N(R)$ with $k$. The key insight for this transition is that the integral in Eq. (17), can be approximated as $\sum_i(logT_i)$ which has $N(R) = k$ terms in it. Here $T_i$ is the distance to the $i^{th}$ neighbor. Now the second and the third term in Eq. (17) can be combined as $\sum_i(logT_k/T_i)$ which leads to the form

$$\frac{k}{m} - \sum_{j=1}^{k} \log \frac{T_k}{T_j} - \gamma \frac{\partial}{\partial m} D(m\|\bar{m}_0) = 0 \quad (18)$$

For the regularization term, taking a first order approximation to a natural log series we can write

$$\frac{\partial}{\partial m} D(m\|\bar{m}_0) = \log \frac{m}{\bar{m}_0} \approx 2\frac{m - \bar{m}_0}{m + \bar{m}_0} \quad (19)$$

This approximation reveals some very interesting facts. Firstly, if we assume, close to optimality, $m + \bar{m}_0 \approx 2\bar{m}_0$, then

$$\frac{\partial}{\partial m} D(m\|\bar{m}_0) \approx \frac{\frac{\partial}{\partial m}(m - \bar{m}_0)^2}{2\bar{m}_0}$$

and the regularization scheme essentially penalizes the variance of the estimator from the current mean estimate within the valid neighborhood. Substituting the approximation from Eq. (19) in Eq (18), we can write the quadratic in $m$ as

$$\left( \sum_{j=1}^{k} \log \frac{T_k}{T_j} + 2\gamma \right) m^2 \quad (20)$$
$$+ \left( \bar{m}_0 \sum_{j=1}^{k} \log \frac{T_k}{T_j} - 2\gamma\bar{m}_0 - k \right) m$$
$$- \bar{m}_0 k = 0$$

The optimization procedure is as follows. We assign random values in the range $[0, d]$ to all $m_i$, $i = \{1, 2, \ldots, n\}$ and fix $\gamma$ to be a small value. For each iteration we fix all other assignments except the $j^{\text{th}}$ and estimate $m_j$, by solving the quadratic mentioned in Eq. (20), based on all the other fixed assignments. As mentioned earlier, the quantity $\bar{m}_0(j,r)$ is obtained by averaging the current estimate at the neighbors of the point $j$ as mentioned in Eq. (14). This process is repeated for all the observations and the estimates are updated. For each subsequent iteration $\gamma \to \max[\gamma(1 + \epsilon), 1]$ where $\epsilon$ is a small positive number. Note that we do not need any further averaging of the estimates for different values of $k$. The stopping condition for the iteration is set as either a maximum number of iterations or the change in the estimates going very close to zero. Finally we estimate the combined intrinsic dimension by computing the mean of the estimate over all the observations.

$$m_{est} = \frac{1}{N} \sum_{i}^{N} m(x_i) \quad (21)$$

For small values of $k$ the estimated ID, from Eq. (20), can be written as

$$m = \frac{k}{\sum_{j=1}^{k} \log \frac{T_k}{T_j} + \gamma \log(m/\bar{m}_0)} \quad (22)$$
$$\approx \frac{k}{\sum_{j=1}^{k} \log \frac{T_k}{T_j}} (1 - \gamma \frac{\log(m/\bar{m}_0)}{\sum_{j=1}^{k} \log \frac{T_k}{T_j}}) \quad (23)$$

The improvement in estimated ID, as compared to estimation with no regularization, can be observed from the Fig.1, which depicts 5D Gaussian points embedded in the same dimension. For small $k$ the estimated ID is almost always greater than the true dimension. For this case the divergence penalty is positive and hence the regularized estimate has lesser positive bias, as evident from the comparison plot. At the junction when the smooth density assumption fails the divergence penalty reverses sign, and helps in reducing the negative bias observed in the unregularized estimator. The overall performance for the regularized method improves over the entire span of the neighborhood size $k$. The variance for the estimates is shown against the sample index. The variance generally improves as $k$ grows. The same color dots correspond to the same value of $k$ for the two methods. The worst variance for the unregularized technique is close to 50, whereas for the regularized technique it is about 13.

### 3.1 Asymptotic Behavior

For the asymptotic behavior we bound the deviation of the estimated $m$ from $m_0$, at a point $x$, and re-write Eq. (18) as

$$\frac{k}{m} - \sum_{j=1}^{k} \log \frac{T_k}{T_j} - \gamma \log \frac{m}{m_0} = 0 \quad (24)$$

Since the rightmost term is independent of $k$, we can condition on $T_k$ and assume the Poisson approximation is exact. Consequently $m \log(T_k/T_j) : 1 \leq j \leq k-1$ are distributed as the order statistics of a sample size of $k-1$ from a standard exponential distribution. Writing the sum of $k-1$ exponentially distributed variables as a Gamma distribution, we get $m \sum_i \log(T_k/T_j) \sim \text{Gamma}(k-1, 1)$. Writing the approximate estimator as a function of $k$ from Eq. (24), for constant $m$, we get

$$\hat{m}_k(x) = \frac{k}{\sum_{j=1}^{k} \log \frac{T_k}{T_j} + \gamma \log \frac{m}{m_0}} \quad (25)$$

The expectation with respect to the Gamma distribution can now be evaluated as

$$E[\hat{m}_k(x)] = \frac{km}{\Gamma(k-1)} \int_0^\infty \frac{1}{x + \gamma \log \frac{m}{m_0}} x^{k-2} e^{-x} dx$$
$$= \frac{km}{\Gamma(k-1)} \int_t^\infty \frac{1}{s}(s-t)^{k-2} e^{-s+t} ds$$
$$= \frac{km}{\Gamma(k-1)} e^t (\int_0^\infty \frac{1}{s}(s-t)^{k-2} e^{-s} ds$$
$$- \int_0^t \frac{1}{s}(s-t)^{k-2} e^{-s} ds)$$

Noting the quantity $t = \gamma \log \frac{m}{m_0} \ll 1$, everywhere in the feasible region of the optimization scheme we can neglect the second term in the parenthesis. This is guaranteed by the fact that for large values of $\log|m/m_0|$, $\gamma$ is very small, and when $\gamma$ becomes large the value of $\log|m/m_0|$ is pushed to be very small. Assuming $k \to \infty$, the mean can be written as

$$E[\hat{m}_k(x)] = me^t(1-t) \approx m(1-t^2) \quad (26)$$

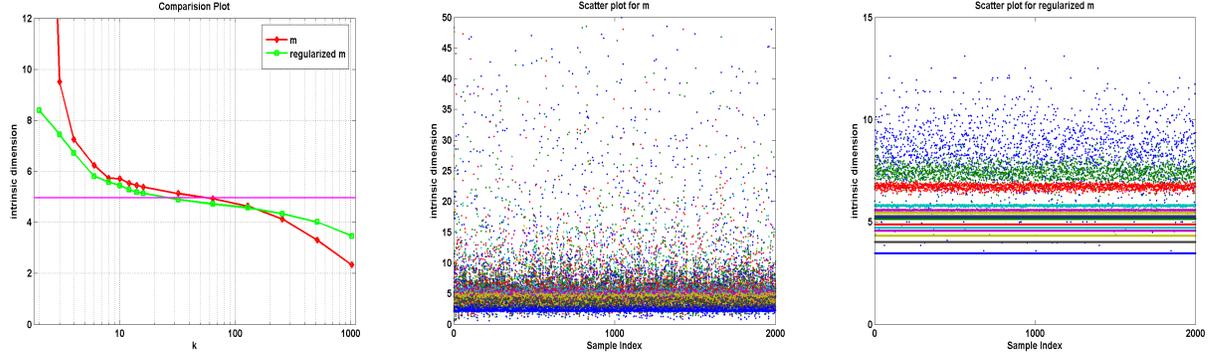

Figure 1: Performance comparison for 5D Gaussian points. Red: $m$ estimator Levina *et al.* [12], green: our method. The pink line denotes the true ID. For the variance plots, the variance improves with $k$ as shown by different colors becoming more and more linear. The same color means the same value of $k$ for both the methods.

Assuming, $E[1/m] \leq M < \infty$, which is also the condition for the maximum likelihood to exist, the variance of the estimator, for a fixed point $x$, can be obtained from the Fisher Informaition criterion,

$$\begin{aligned}
I(\hat{m}_k(x)) &= -E[\frac{\partial^2}{\partial m^2} \log L_R(m)] \\
&= \frac{k}{m^2} + \gamma E[1/m] \\
0 < I_n(m) < I_n(\hat{m}_r) &\leq I_n(m) + \gamma M \\
1/I_n(m) > 1/I_n(\hat{m}_r) &\geq 1/(I_n(m) + \gamma M) \\
var(m) > var(\hat{m}_r) &\geq \frac{var(m)}{1 + \frac{\gamma M}{2I_n(m)}}
\end{aligned}$$

where $I_n(m) = \frac{k}{m^2}$ is the Fisher information for the unregularized MLE. The improvement in the asymptotic variance is guaranteed by the form of the regularizer, as mentioned earlier. The development till this point assumes the directed form of the KL divergence mentioned in Eq. (8). If we assume the reverse direction for the KL divergence, then the regularizer essentially penalizes the bias of the estimator, but increases the variance. The main drawback with the reverse divergence is the estimator of the term $m_0$, whose ML estimate based on the current estimates at the neighboring points can be shown to be the *geometric mean* of the neighborhood estimates. It remains a future direction to penalize with respect to the symmetric KL divergence.

## 4 Results

To establish the acceptability of our method, as a possible ID estimation technique, we generate comparisons with the method proposed by Levina and Bickel [12]. Their method was claimed to perform better overall then the correlation dimension based approach mentioned in [7, 2] and a direct regression based approach based on regressing $\log T_k$ to $\log k$. We identified that since [12] already compared their method to the existing techniques, we select the inverse $m$ estimator proposed by MacKay and Ghahramani [13] as the third method against which we perform comparative simulations. For the first set of experiments we select the 1D datasets mentioned in [26], embedded in 2 and 3D. Fig. 2 shows the results for these experiments. The three scatter plots, in each row, next to each example show the estimator variance with respect to increasing number of neighbors ($k$) for each method. Generaly the dot patterns become more linear with increasing $k$. The same color of dots, correspond to the same number of neighbors ($k$) across the three methods. The horizontal axis for the plots run over the sample indices. Note the spread of values for our method is the least compared to both the other unregularized techniques for many $k$'s. The fourth column shows the mean estimate for each value of $k$. Note the better bias properties for our estimator (green) for a wider range of values of $k$. Also note the true dimension is marked with a pink line. For the curve with a singular point (bottom row), note the graceful deterioration of the performance of the proposed regularized technique (green curve) over the other two techniques. In Fig. 3, we present result for a composite manifold for which our method again outperforms both the other techniques.

Next we present results for two standard datasets, S-curve data and the swiss-roll data in Fig. 4. These datasets are intrinsically 2D embedded into 3D. So $m = 2$ and $d = 3$. We generate the ML estimate of $m$ as a function of $k$ and plot the results. The scatter plots are spread out more than the earlier plots due to the irregular sampling as well as higher amount of noise present in the datasets. Our method clearly outperforms the other two techniques, both in terms of variance (scatter plots) for the entire range of $k$. The bias performance is best for [13] for very small $k$, but our method quickly catches up. Note the correct dimension is again marked with a pink line across the plots in the fifth column.

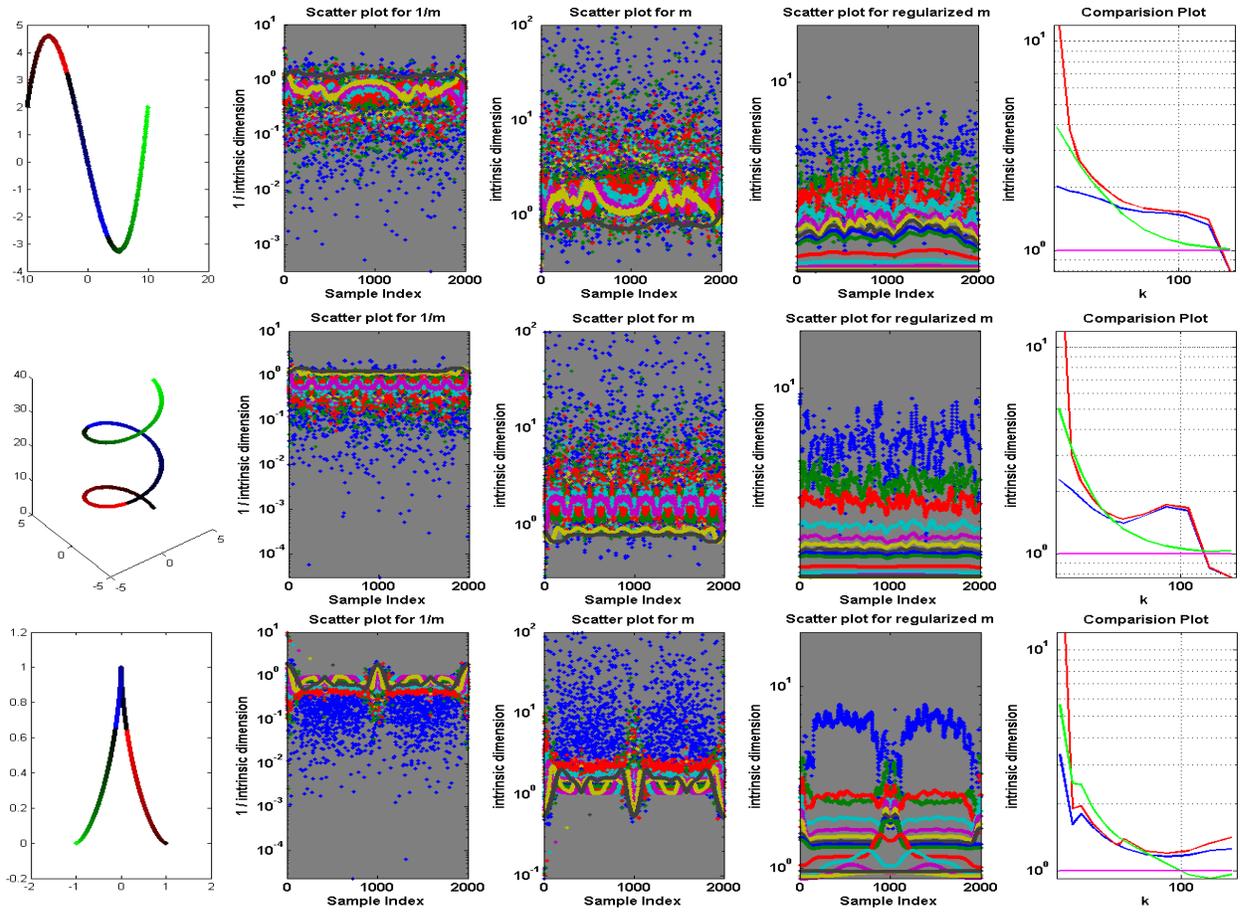

Figure 2: Left to right: First column (top to bottom): 1D curves embedded in 2D and 3D. Second to fourth columns: scatter plots for $1/m$ estimator Mackay *et al.* [13], $m$ estimator Levina *et al.* [12], and our method. Fifth column: comparative curves blue curve [13], red curve [12], green curve our method. Pink line shows the true dimension. Note the last example has a singular point.

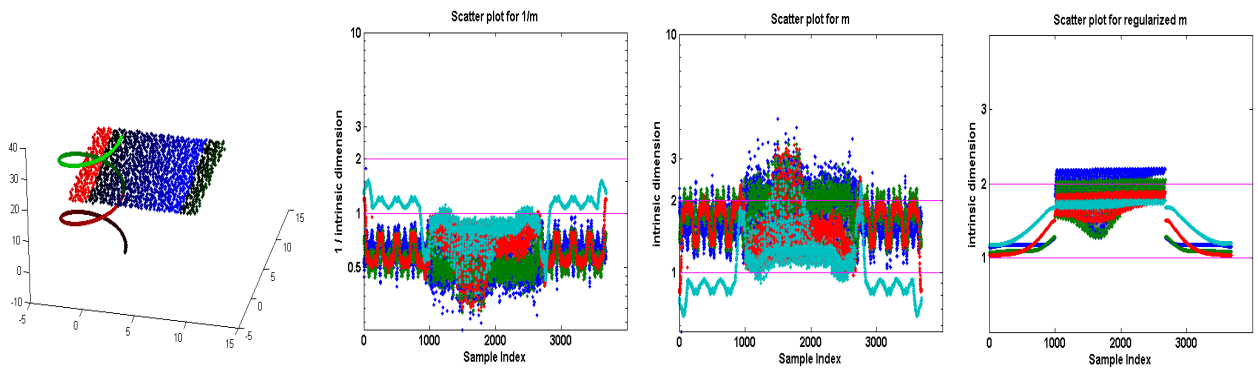

Figure 3: Composite manifold. Second to fourth columns: scatter plots for $1/m$ estimator Mackay *et al.* [13] (underestimates ID), $m$ estimator Levina *et al.* [12] (overestimates ID), and our method estimates the two dimensions correctly, away from the flux regions. Pink lines represent the true dimensions.

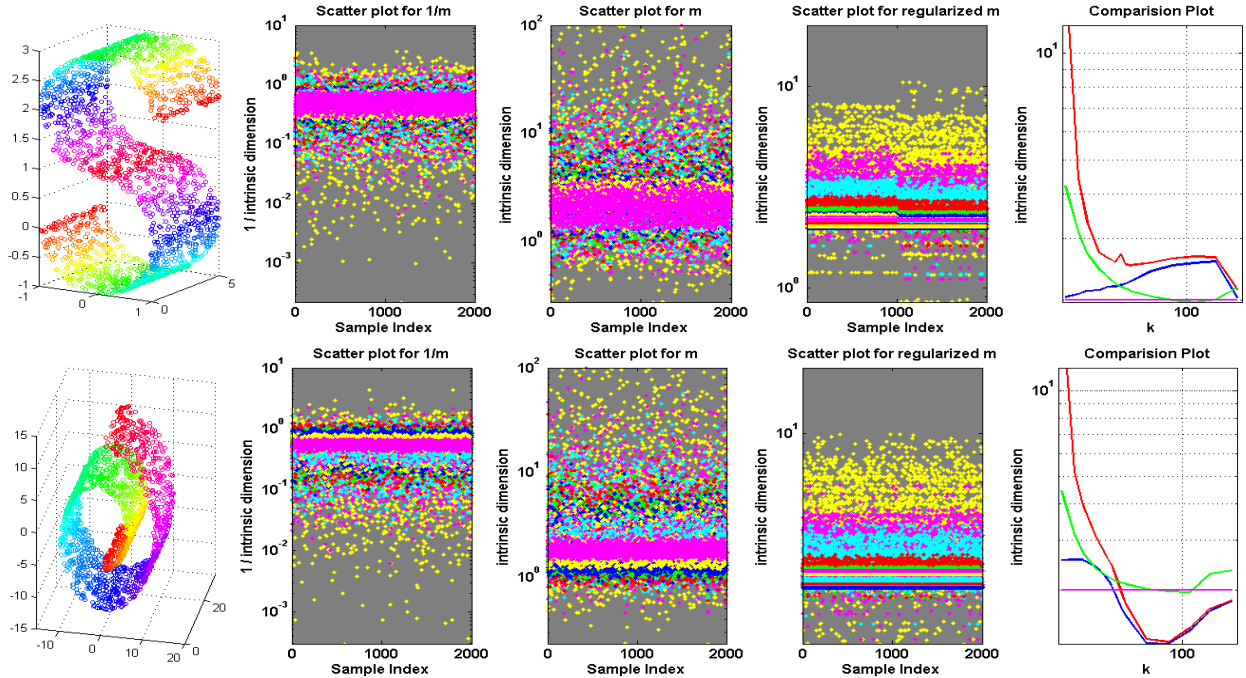

Figure 4: S-curve and Swissroll. Blue curve $1/m$ estimator [13], red curve $m$ estimator [12], green curve our method. Pink line shows the true dimension.

Finally we look into the two real datasets which have found lot of attention within the dimensionality reduction as well as ID estimation community. The two datasets are face dataset from the ISOMAP work [24] and the rotating cup in hand CMU dataset[2]. For the face dataset, there are 698 images of human faces generated by using three free parameters: vertical and horizontal orientation, and light direction. Similarly for the rotating cup dataset there are 481 images. The image sizes are $64 \times 64$, which make $d = 4096$. It has been argued that the expected dimension for the faces should be 3 since the only three variable are the two orientations and illumination direction. The number of data points is roughly $10\%$ and $15\%$ of the dimension for cup and faces respectively, and so very accurate estimate is not possible. But the results obtained by all the three methods are quite similar to each other for the range of $k$ denoted in Fig. 5. For the hand dataset our estimator is closest to the value 3, and for the face dataset our method performs very close to the $1/m$ estimator. This suggests that our method is right there with the competing schemes as far as the estimation accuracy is concerned, and might be the first to converge to the actual estimate given enough data points.

## 5 Conclusion

We have presented a new algorithm to estimate the intrinsic dimension of data sets. We draw motivation from divergence minimization principles to penalize the maximum likelihood estimator and improve the variance performance of the estimator. The results show improvement in bias performance for not very small neighborhoods. This two fold improvement can be explained by the fact that the unconstrained method greatly underestimated the bias for larger neighborhoods and overestimated for smaller neighborhoods. By careful penalty switching we provide a better mean estimate over a wider range of neighborhood sizes. The weighting term $\gamma$ in our scheme switches the estimation of the ID to a purely divergence minimization scheme which performs better at higher values of $k$. We present comparative results on noisy, synthetic as well as real data. The results validate our claims that the technique can be used in real scenarios to generate a valid estimate of the intrinsic dimension.

---

[2]http://vasc.ri.cmu.edu/idb/html/motion/hand/index.html

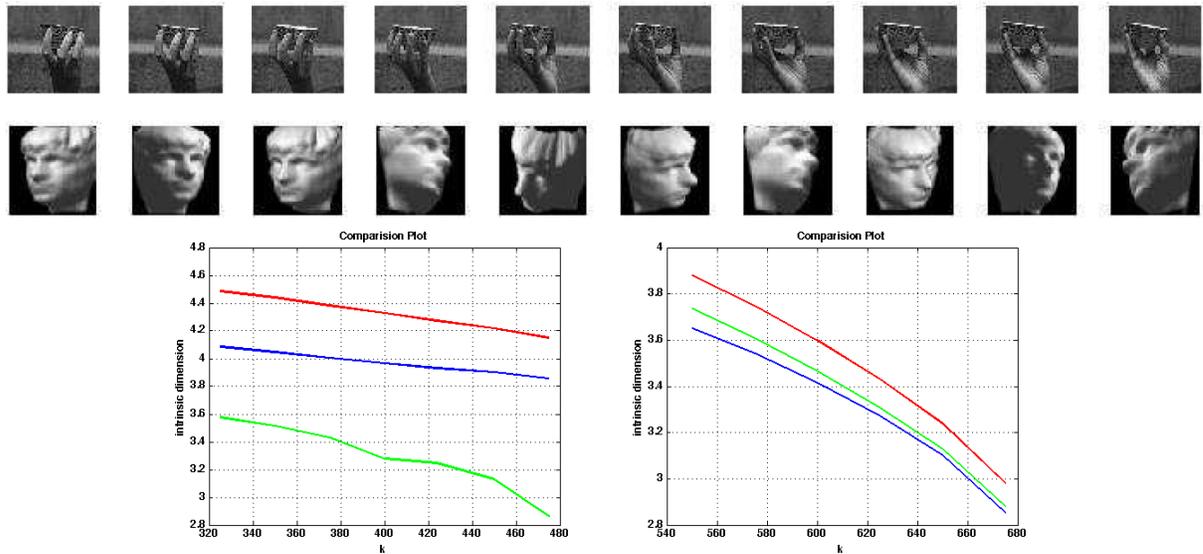

Figure 5: Sample images from the Hands and Face datasets. Comparative plots for hand (left) and faces (right). Red curve [12], blue curve [13], green curve our method.